\pdfoutput=1

\documentclass[11pt]{article}

\usepackage[]{acl}

\usepackage{times}
\usepackage{latexsym}

\usepackage[T1]{fontenc}

\usepackage[utf8]{inputenc}

\usepackage{microtype}

\usepackage{inconsolata}

\usepackage{graphicx}

\usepackage{subcaption}
\usepackage{amsmath}
\usepackage{tabularx}
\usepackage{ulem}
\usepackage{multirow}
\usepackage{array}
\usepackage{enumitem}

\title{Mixture of Diverse Size Experts}

\author{Manxi Sun, Wei Liu, Jian Luan, Pengzhi Gao, and Bin Wang \\
Xiaomi AI Lab, Beijing, China \\
\texttt{\{sunmanxi, liuwei40, luanjian, gaopengzhi, wangbin11\}@xiaomi.com}
}

\begin{document}

\maketitle

\begin{abstract}

The Sparsely-Activated Mixture-of-Experts (MoE) has gained increasing popularity for scaling up large language models (LLMs) without exploding computational costs. Despite its success, the current design faces a challenge where all experts have the same size, limiting the ability of tokens to choose the experts with the most appropriate size for generating the next token. In this paper, we propose the Mixture of Diverse Size Experts (MoDSE), a new MoE architecture with layers designed to have experts of different sizes. Our analysis of difficult token generation tasks shows that experts of various sizes achieve better predictions, and the routing path of the experts tends to be stable after a training period. However, having experts of diverse sizes can lead to uneven workload distribution. To tackle this limitation, we introduce an expert-pair allocation strategy to evenly distribute the workload across multiple GPUs. Comprehensive evaluations across multiple benchmarks demonstrate the effectiveness of MoDSE, as it outperforms existing MoEs by allocating the parameter budget to experts adaptively while maintaining the same total parameter size and the number of experts.

\end{abstract}

\section{Introduction}

Large Language Models (LLMs) have demonstrated remarkable performance in a variety of NLP tasks and have become valuable assistants through a wide range of applications. The scaling law \cite{kaplan2020scaling} demonstrates that larger models exhibit superior performance. However, training larger models requires increased computational resources, posing a critical challenge. Mixture-of-Experts (MoE) \cite{10.5555/3586589.3586709,lepikhin2021gshard} address this challenge by using sparse activation to scale up the trainable parameters while maintaining high training and inference efficiency. Recent MoE-based architectures, such as Mixtral of Experts \cite{jiang2024mixtralexperts}, DeepSeekMoE \cite{dai-etal-2024-deepseekmoe}, and OpenMoE \cite{xue2024openmoeearlyeffortopen} have shown superior performance in various tasks. 

Specifically, \citet{dai-etal-2024-deepseekmoe} discuss two main issues in the design of the MoE Feed-Forward Networks (FFNs) architecture: Knowledge Hybridity, where each expert covers diverse knowledge due to the limited number of experts, and Knowledge Redundancy, where multiple experts share common knowledge. To address these issues, they propose Fine-Grained Expert Segmentation by splitting the FFN intermediate hidden dimension and Shared Expert Isolation by isolating certain experts to be always activated as shared experts. Additionally, \citet{zhao2024hypermoebettermixtureexperts} introduce Hypernetworks and HyperExperts modules to capture the cross-expert and cross-layer knowledge. 

However, almost all existing MoE architectures consist of experts with identical structures and sizes. This homogeneous architecture becomes a significant bottleneck when generating tokens with varying difficulty; some tokens are easier to predict, while others are more challenging. To deal with the varied difficulty, we propose the Diverse Size Experts structure for each FFN layer, where each expert has a different parameter size to handle generating tasks of varying difficulty. Note that we find a similar recent work called Heterogeneous the Mixture of Experts \cite{wang2024hmoeheterogeneousmixtureexperts}, which shares a similar motivation and utilizes parameter penalty loss and router entropy loss to control the size and number of activated experts. 

Our contributions are summarized as follows:

\begin{itemize}
\item \textbf{Diverse Size Experts } We introduce the Mixture of Diverse Size Experts (MoDSE) in Section \ref{3}, a new type of FFN layer designed for the MoE framework. Unlike conventional MoEs, which consist of experts of the same size, MoDSE has experts of different sizes. It assigns each token to the expert that best matches its prediction needs in terms of capability, thereby enhancing the model's ability.
    
\item \textbf{Load Balance } GPU nodes containing larger experts in MoDSE will have a heavier workload. To address this issue, we propose the expert-pair allocation method in Section \ref{3.2}, which ensures that each GPU node carries an even distribution of parameters, thus maintaining load balance.
    
\item \textbf{Empirical Validation } MoDSE outperforms conventional MoE with a lower loss value across a diverse set of benchmarks in the $700M \times 8$ model setting, confirming the effectiveness of our approach. We present the evaluation results in Section \ref{4.2}.
    
\item \textbf{Token Routing Analysis } We collect the routing distribution of tokens in both the MoE baseline model and MoDSE, and conduct a thorough analysis in Section \ref{Analysis on Token Routing}. MoDSE exhibits an equally even distribution as the baseline. Additionally, we analyze tokens that are more difficult to predict and find that they are better predicted when routed to an expert which is better suited to handle them.
\end{itemize}

\begin{figure*}[ht]
\centering
\includegraphics[width=0.7 \linewidth]{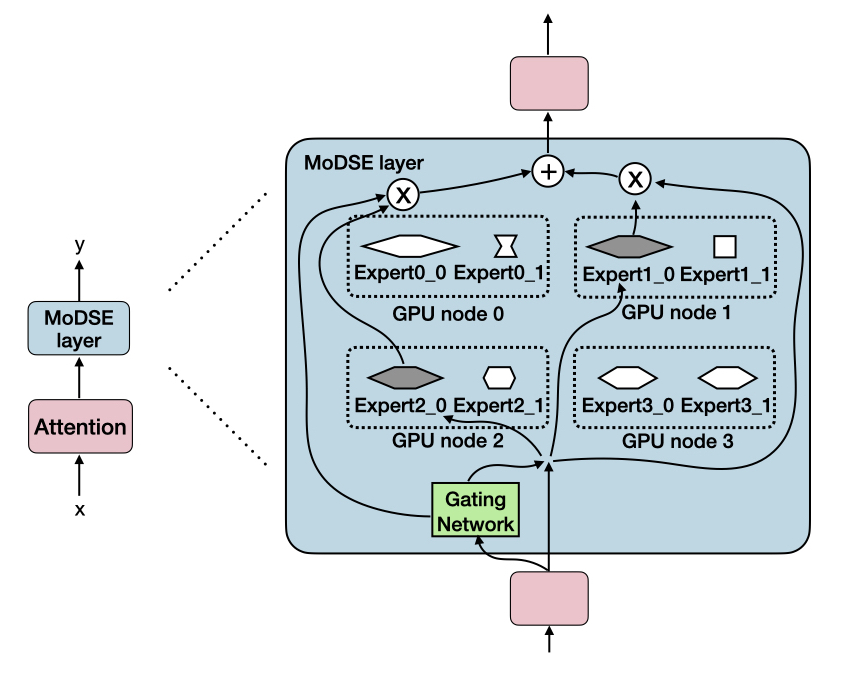}
\caption{Overview of a MoDSE layer with different sizes of experts. In this case, expert1\_0 and expert2\_0 are selected. With the output of the gating network, the outputs of two experts are integrated.}
\label{MoDSE}
\end{figure*}

\section{Preliminaries: Mixture of Experts}

MoE models are usually constructed by replacing dense FFNs layers in the Transformer \cite{NIPS2017_3f5ee243} with MoE layers. An MoE layer typically consists of multiple experts $E_1(\cdot) \cdots E_N(\cdot)$ and the corresponding gate model $G_1(\cdot) \cdots G_N(\cdot)$, $N$ indicates the number of the experts. The gate model \cite{shazeer2017s} with trainable weight matrices $W_g \in \mathcal{R}^{h_{input} \times h}$ and $W_n \in \mathcal{R}^{h_{input} \times h}$ selects the top $k$ experts and combines the outputs of experts to produce the output $y \in \mathcal{R}^h$, where $h_{input}$ is the dimension of input $x$ and $h$ is the dimension of the hidden layer. \citet{10.5555/3586589.3586709} set $k$ as one, while \citet{lepikhin2021gshard,jiang2024mixtralexperts} set as two. The outputs of experts are added with the noise to help with load balance. The noise generated from the input hidden vector $x$ is multiplied by $W_n$ and processed by $Softplus$ and the Root Mean Square Layer Normalization function (RMSNorm), where $\gamma$ is a learnable coefficient.

\begin{equation}
y=\sum_{i=1}^N G_i(x)E_i(x)
\end{equation}\label{moeeq}
\begin{equation}
G(x) = Softmax(\text{KeepTopK}(H(x),k))
\end{equation}
\begin{equation}\nonumber
H(x)_i=(x \cdot W_g)_i + \text{RMSNorm}(f((x \cdot W_{n})_i)) 
\end{equation}
\begin{equation}
\text{KeepTopK}(v,k)_i =\left\{ 
\begin{array}{ll}
    v_i&  v_i \in \text{topk(v)}\\
    -\infty & \text{otherwise } 
\end{array}
\right.
\end{equation}
\begin{equation}
\text{RMSNorm}(x) = \gamma \cdot \frac{x}{\sqrt{\frac{1}{n} \sum_{i=1}^{n} x_i^2 + \epsilon}}
\end{equation}
\begin{equation}
f(\cdot) = Softplus(\cdot) = \log (1+\exp (\cdot))
\end{equation}

\section{MoDSE Architecture}\label{3}

Predicting the next token is easier within frequently appearing token pairs in the corpus. Tokens within the same word or phrase are easier to generate than those between two phrases or words. Analogous to the human brain, the amount of thought required to generate the next word varies among different words. Inspired by the fact that the difficulty of generating each next token varies, we propose MoDSE as shown in Figure \ref{MoDSE}. In our work, the size of the expert parameters is used to quantify the amount of thinking involved. We assign experts a range of parameter sizes by setting the dimensions of the hidden layers to various lengths. However, the imbalance in expert size leads to an uneven workload. To address this issue, we propose a meticulously designed expert-pair allocation method to ensure each GPU node's workload is evenly distributed.

\subsection{Diverse Size Experts}

In a traditional MoE structure \cite{10.5555/3586589.3586709,lepikhin2021gshard}, the gating network combines a set of experts with the same size. We here adjust the scale of experts to ensure that different experts can handle tasks of varying difficulty. Note that we denote the designed Diverse Size Experts as $\{\hat{E}_1(\cdot), \cdots, \hat{E}_N(\cdot) \}$, and the dimension of the hidden layer for $\hat{E}_i(\cdot) $ is $\hat{h}_i$. 
\begin{equation}
\hat{y}=\sum_{i=1}^N \hat{G_i}(x)\hat{E}_i(x) 
\end{equation}
\begin{equation}
{(i_1^1, i_1^2),\cdots,(i_n^1, i_n^2)} \text{, with } n=\frac{N}{2}
\end{equation}
\begin{equation}
\hat{h}_{i_k^1} + \hat{h}_{i_k^2} = 2 \times h, \text{, with } k \in 1 \cdots n
\end{equation}
\textbf{To maintain the overall parameter size}, the experts are grouped into pairs $(i_k^1, i_k^2)$, where $k \in 1 \cdots n$ indicates the pair of the experts. The average value of $\hat{h}_i$ within each pair equals $h$, with one expert being larger than the average size and the other smaller. Typically, the number of experts is even, ensuring the experts can be grouped into pairs, thus the total parameter size of the MoDSE model matches that of the vanilla MoE model.

\subsection{Load Balance Consideration} \label{3.2}

In MoDSE, experts with hidden layer sizes larger than the average have a higher workload due to the increased number of parameters, both during training and inference phrases. To address this load imbalance problem, we propose the \textbf{expert-pair allocation} strategy, which places each pair of experts on the same GPU and ensures that each GPU contains an equal number of parameters. For instance, in Figure \ref{MoDSE}, expert pairs are enclosed by dotted line frames, with expert 0 and expert 1 on the same GPU, and so forth.

Besides the standard cross entropy (CE) loss, we use the auxiliary load balance loss $L_a$ from Switch Transformers \cite{10.5555/3586589.3586709} to penalize the unbalanced routing distribution among experts. Consequently, each expert has the same frequency of being routed. In Section \ref{Analysis on Token Routing}, we will demonstrate that after the entire training process, all tokens in the pre-training dataset are evenly spread across all experts. Along with the expert-pair allocation method, this ensures that the final workload of each GPU is balanced. 
\begin{equation}
L_a = \alpha \cdot N \cdot \sum_{i=1}^N f_i \cdot P_i,
\label{auxiliary loss}
\end{equation}
where $\alpha$ is a scalar hyperparameter. $f_i$ is the fraction of tokens routed to expert $i$, $i \in \{1,2, \cdots, N\}$:
\begin{equation}
f_i=\frac{1}{T} \sum_{x \in \text{Batch}} \textbf{\textsc{1}}\{\text{argmax } p(x) =i\},
\label{auxiliary fraction}
\end{equation}
\begin{equation}
p(x) = [p_1(x), p_2(x), \cdots, p_N(x)],
\end{equation}
where $T$ is the number of tokens and $P_i$ is the fraction of the router probability for expert $i$:
\begin{equation}
P_i=\frac{1}{T} \sum_{x \in \text{Batch}} p_i(x)
\label{auxiliary prob}
\end{equation}

\section{Experiments}

\subsection{Experimental Setup}

\paragraph{Models}

Our baseline MoE structure is based on the Llama 2 model \cite{touvron2023llama2openfoundation} with the dense FFNs layers replaced by expert layers. Table \ref{model arch} summarises the model architecture parameters. For the MoDSE setting, we adjust the expert sizes in baseline by modifying the dimensions of the hidden layers in $300M \times 8$ and $700M \times 8$ settings, as listed in Table \ref{MoDSE dims}. There are 8 experts grouped into 4 pairs, with the ratio to the input size as $(4.5, 0.5)$, $(4.0, 1.0)$, $(3.0, 2.0)$, and $(2.5, 2.5)$. We train byte pair encoding (BPE)  \cite{sennrich2016neuralmachinetranslationrare} tokenizer with both English and Chinese datasets, and use it in the following experiments.

\begin{table}[ht]
\centering
\begin{tabular}{c | c c}
\hline
\textbf{Parameter} & \textbf{$300M \times 8$} & \textbf{$700M \times 8$} \\
\hline
dim & 1536 & 2048\\
n\_layers & 8 & 12\\
\# heads & 12 & 32\\
\# expert & 8 & 8\\
top $k$ & 2 & 2 \\
vocal\_size & 30064 & 30064\\
h & 3840 & 5120 \\
\hline
\end{tabular}
\caption{MoE model architecture with $300M \times 8$ and $700M \times 8$ parameters, both with identical expert sizes.}
\label{model arch}
\end{table}

\begin{table}[ht]
\centering
\begin{tabular}{c c}
\hline
\textbf{Model} & \textbf{Expert size pairs}  \\
\hline
$300M \times 8$ & [(6912,768), (6144,1536),\\
 & (4608,3072), (3840,3840)]\\
 \hline
$700M \times 8$ & [(9216,1024), (8192,2048),\\
& (6144,4096), (5120,5120)]\\
\hline
\end{tabular}
\caption{The list of expert pair sizes in $300M \times 8$ and $700M \times 8$ parameters.}
\label{MoDSE dims}
\end{table}

\paragraph{Training configurations}
We utilize the Adam optimizer \cite{Kingma2014AdamAM}, with hyperparameters $\beta_1 = 0.9$, $\beta_2 =0.95$, $\text{eps} = 1e-8$, $\text{weight decay} = 0.1$ and $\text{gradient clipping} = 1.0$. We use a cosine learning rate schedule \cite{loshchilov2017sgdr}, such that the initial learning rate is 2e-7, the warm-up update steps are 2000 and the minimal learning rate is 3e-5. We employ the ZeRO optimization \cite{rajbhandari2020zeromemoryoptimizationstraining} for distributed training. All experiments are carried out on clusters equipped with NVIDIA A800 GPUs. The A800 cluster features 8 GPUs per node, interconnected using NVLink and NVSwitch within nodes. Two nodes are used for the $300M \times 8$ setting, and 8 nodes are used for the $700M \times 8$ setting.

\paragraph{Datasets}
We collected 100B tokens training data from various reputable sources for pre-training. This dataset includes both English and Chinese language, and spans multiple fields, including CommonCrawl \cite{wenzek-etal-2020-ccnet}, code, academic papers, books, mathematics, and Q\&A.

\subsection{Main Results} \label{4.2}
\textbf{Evaluations}
We evaluate models in downstream tasks using in-context learning including AGIEval \cite{zhong2023agievalhumancentricbenchmarkevaluating}, MMLU \cite{hendrycks2021measuringmassivemultitasklanguage}, GSM8K \cite{cobbe2021trainingverifierssolvemath}, LAMBADA \cite{paperno2016lambadadatasetwordprediction}, MATH \cite{hendrycks2021measuringmathematicalproblemsolving}, TriviaQA \cite{joshi2017triviaqalargescaledistantly}, PIQA \cite{Bisk2019PIQARA}, SIQA \cite{sap-etal-2019-social}, and INTENT from \citet{XIAOAI} which contains 43 different user intention classes. The model with identical expert sizes is used as the baseline, all the evaluation results are listed in Table \ref{eval}.

\begin{table}[ht]
\centering
\begin{tabular}{c | c c}
\hline
\textbf{Benchmark} & \textbf{Baseline} & \textbf{MoDSE} \\
\hline
AGIEval (Acc.) & 26.2 & \textbf{28.1}\\
MMLU (Acc.) & 26.5 & \textbf{29.9}\\
INTENT (Acc.) & 13.6 & \textbf{16.5}\\
GSM8K (EM) & 5.9 & \textbf{7.7}\\
LAMBADA (EM) & 36.8 & \textbf{38.9}\\
MATH (EM) & 0.8 & \textbf{2.6}\\
TriviaQA (EM) & 5.2 & \textbf{8.3}\\
PIQA (EM) & 53.1 & \textbf{57.6}\\
SIQA (EM) & 42.9 & \textbf{60.9}\\
\hline
\end{tabular}
\caption{Comparison between MoE baseline and MoDSE on size of $700M \times 8$. The bold font indicates the better. With the same parameter, MoDSE achieves better performance than the baseline. All the tasks are fewshot in context learning, and GSM8k includes 8 shots examples and others include 5 shots examples.
}
\label{eval}
\end{table}

\begin{figure*}[ht]
\centering
\includegraphics[width=1.0 \linewidth]{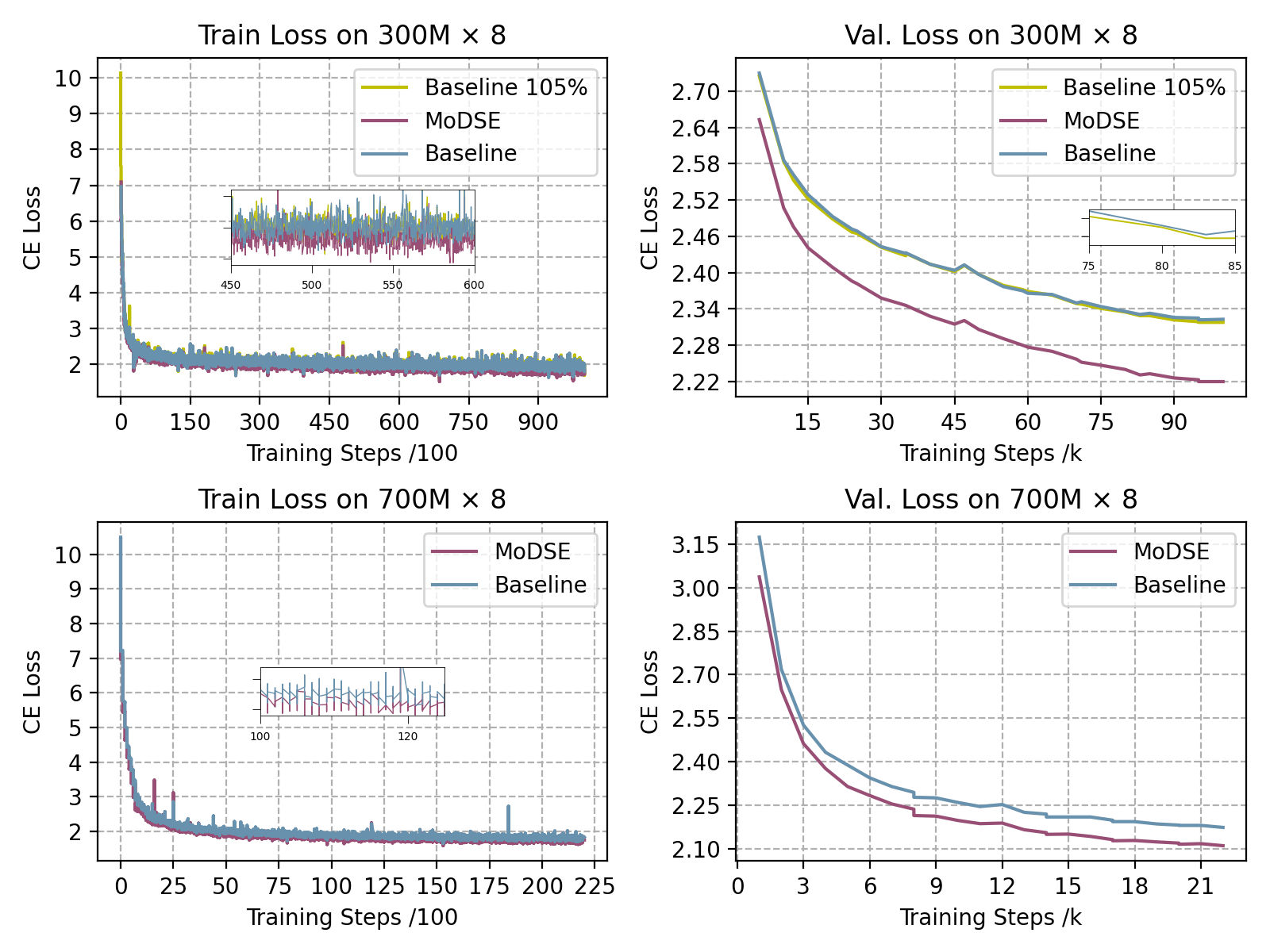}
\caption{Training and validation loss curves for the $300M \times 8$ and $700M \times 8$ models, with cross-entropy loss values indicated on the curves.}
\label{train and vad}
\end{figure*}

\paragraph{Training convergence} 
As shown in Figure \ref{train and vad}, in both $300M \times 8$ and $700M \times 8$ settings, MoSDE demonstrates an earlier convergence and exhibits a lower cross-entropy loss value, compared to the baseline throughout the entire training process in both 300M × 8 and 700M × 8 settings. Additionally, MoSDE on both settings outperforms the baseline on the validation set.

We use the average hidden size of the chosen experts with 10B tokens data to represent the model workload. The experiment shows that the average workload across experts in MoSDE is 4045, compared to 3840 in the baseline.
We also conduct experiments on 4045 as the same size as MoSDE, which corresponds to Baseline 105\% in Figure \ref{train and vad}. The result shows that the curve of Baseline 105\% aligns closely with the original baseline. Thus, MoSDE shows better convergence features even without the benefits of a slightly larger workload capacity. We analyze the benefits of distinct expert sizes in the following section.

\begin{figure*}[ht]
\centering
\includegraphics[width=1.0 \linewidth]{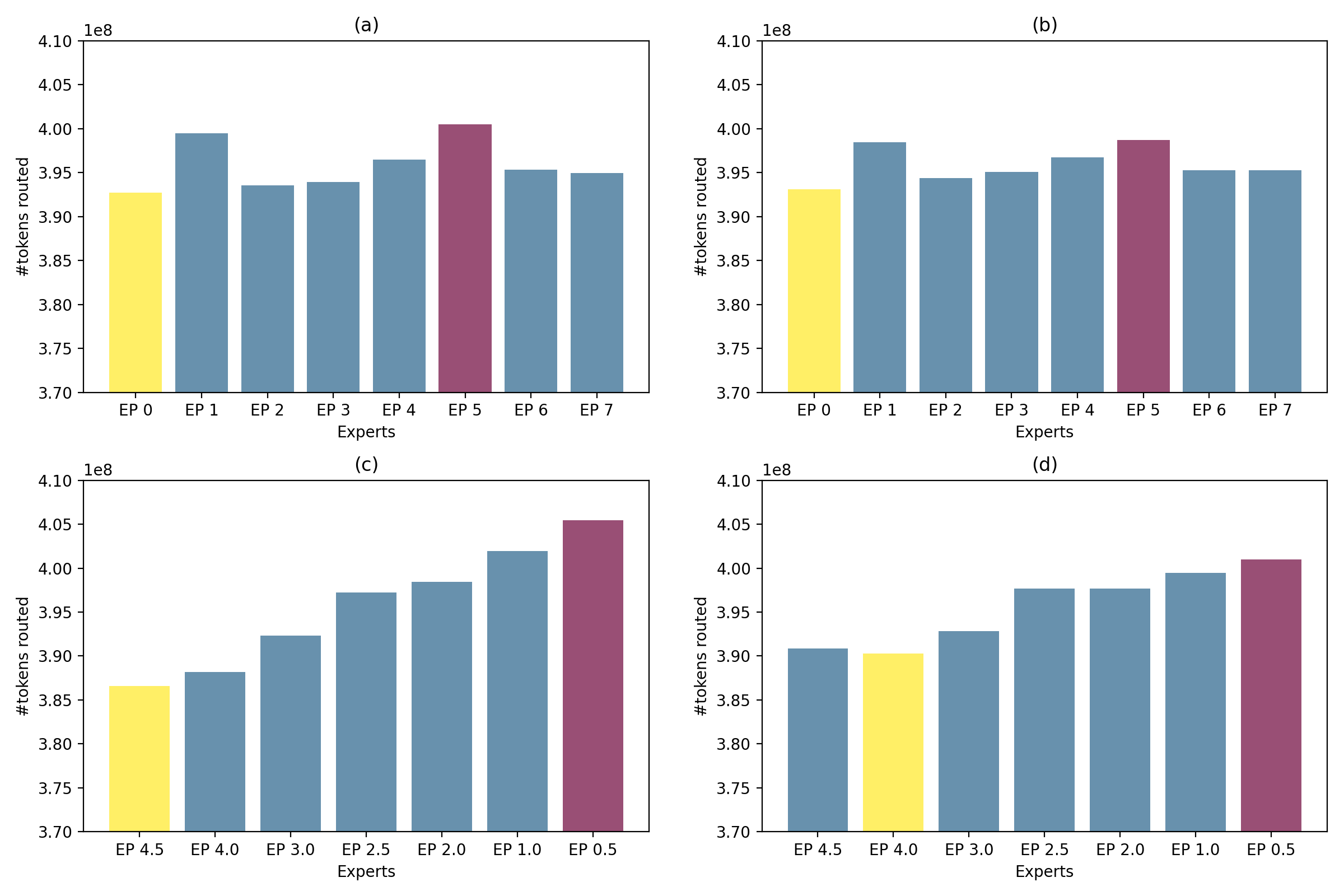}
\caption{The number of tokens routed to each expert. The bar is the sum of the number across the layers. Figure (a) shows results in Baseline in epoch 2, and (b) in the last epoch. Figure (c) shows results in MoDSE in epoch 2, and (d) in the last epoch. The purple bar indicates the most routed expert, and the yellow indicates the least.}
\label{epdist}
\end{figure*}

During training, the distribution of tokens routed to each expert in MoDSE becomes more and more balanced. As illustrated in Figure \ref{epdist}, the distribution is initially uneven, with smaller experts receiving more tokens and the largest expert receiving the fewest, as shown in Figure \ref{epdist} (c). By the end of the training process, as depicted in Figure \ref{epdist} (d), the distribution evens out, ensuring the workload balance described in Section \ref{3.2}.

\paragraph{Decoding efficiency}
We record the inference duration for both the baseline and MoDSE models during the evaluation of downstream tasks, with the results presented in Table \ref{speed}. The inference times for both models are quite comparable. As illustrated in Figure \ref{epdist} (d) and the final table in Table \ref{MoDSE token routing}, tokens are nearly uniformly distributed across different experts in the last training epoch. Given the expert-pair allocation strategy described in Section \ref{3.2}, the inference speeds of the baseline and MoDSE models should be similar.

\begin{table}[ht]
\centering
\begin{tabular}{c | c c c}
\hline
\textbf{Benchmark} & \textbf{MoE} & \textbf{MoDSE} \\
\hline
AGIEval & 48s & 59s  \\
MMLU & 3min 26s & 3min 27s \\
INTENT & 1min 31s & 1min 34s \\
GSM8K & 20min 26s & 20min 43s \\
LAMBADA & 40min44s & 40min48s \\
MATH & 21min 21s & 21min 34s \\
TriviaQA & 46min 53s & 48min 55s \\
PIQA & 44min56s & 43min34s  \\
SIQA & 2min35s & 2min36s  \\
\hline
\end{tabular}
\caption{The inference duration of the baseline and MoDSE models on downstream tasks. The AGIEval task contains 615 examples, the MMLU task contains 2341 examples, the INTENT task contains 741 examples and the rest tasks with 100 examples.
}
\label{speed}
\end{table}

\subsection{Analysis on Token Routing } \label{Analysis on Token Routing}
We further conduct the experiments on 10B tokens data, to analyze the choices of tokens. 
The statistics from the 2nd to the 7th epoch are listed in Appendix \ref{token routed}.
The baseline model shows an even distribution of experts' workload. The ratio between the largest and the smallest number of tokens routed to the experts ranges from 1.2 to 3.0. The statistics for the MoDSE setting show a non-uniform distribution, with ratios larger than 3.0 appearing, particularly in the first 2 layers of the model and for the experts with the second largest probability. 

However, after the entire training process, in the last epoch, only one ratio remains larger than 3.0, with the others ranging from 1.5 to 3.0, indicating that the token distribution among experts becomes more balanced by the end of the training. 

As shown in Figure \ref{epdist} (c, d) and Table \ref{MoDSE token routing}, it is notable that the experts chosen by the most tokens are not always the ones with larger sizes. Conversely, experts with larger sizes can sometimes be the least visited by the tokens.

\paragraph{Analysis on Difficult Tokens}
We track the tokens in the MoDSE setting which exhibits a higher cross entropy (CE) loss than the mean value of 1.05 in the baseline, considering them having greater prediction difficulty. The average CE loss values in the MoDSE setting are lower than those in the baseline, indicating that MoDSE improves generating ability. This improvement is achieved by routing tokens that are more difficult to predict to the expert whose size better fits the token's generating task. Table \ref{dtoken loss dist} shows the results for the tokens with a higher CE loss than the mean loss value. The tokens in the higher loss threshold show a larger loss decline in the MoDSE setting, demonstrating that the MoDSE model performs better on more difficult tokens.

\begin{table}[ht]
\centering
\begin{tabular}{c | c c}
\hline
\textbf{loss threshold} & \textbf{avg. loss red.}& \textbf{\#tokens}  \\
\hline
2.0 & 0.58 & 180\\
1.8 & 0.46 & 222 \\
1.6 & 0.36 & 337 \\
1.4 & 0.32 & 730 \\
1.2 & 0.22 & 1991 \\
1.05 & 0.18 & 3633 \\

\hline
\end{tabular}
\caption{Average CE loss reduction across different intervals. The higher the initial CE loss, the more significant the improvement demonstrated by the MoDSE model. The avg. loss red. stands for the average CE loss decrease from baseline to MoDSE.}
\label{dtoken loss dist}
\end{table}

\paragraph{Difficult Tokens Routing Distribution}
To identify which experts handle the difficult tokens, further analysis is conducted on the 180 tokens with a CE loss greater than 2.0 in the baseline setting. We track the distribution of these 180 difficult tokens across the distinct experts in the 10B tokens data using the converged training model checkpoint. The full tracking results can be found in Appendix \ref{token dist on de}. 

\begin{figure}[ht]
\centering
\includegraphics[width=1.0 \linewidth]{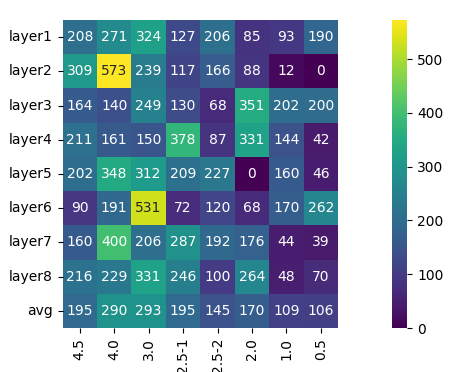}
\caption{The top one expert choice of difficult tokens across eight layers. More tokens are routed to larger experts, distributed on the left half of the heat map.}
    \label{heatmap}
\end{figure}

For these difficult tokens, as shown in Figure \ref{heatmap} and Table \ref{token dist across ex}, more tokens choose the larger experts, while fewer tokens select the smaller experts. This phenomenon is even more pronounced when only considering the top one expert. More than twice as many tokens (6215) chose the larger experts compared to the smaller ones (3085). This result indicates that the larger experts, with capabilities to handle tokens with more difficult prediction tasks, are more frequently chosen by tokens facing more challenging next-token generation tasks.

\begin{table}[ht]
\centering
\begin{tabular}{c|c c}
\hline
\textbf{Expert Size} & \textbf{\#tokens to} & \textbf{\#tokens to}\\
& \textbf{top1 \& 2} & \textbf{top1}\\
\hline
4.5 & 2649 & 1560\\
4.0 & 3729 & 2313\\
3.0 & 4095 & 2342\\
2.5 & 2332 & 1166\\
2.5 & 2933 & 1566\\
2.0 & 2877 & 1363\\
1.0 & 2972 & 873\\
0.5 & 2477 & 849\\
\hline
sum(L) & 10473 & 6215 \\
sum(S) &8326 & 3085\\
\hline
\end{tabular}
\caption{The distribution of difficult tokens across different experts. The sum(L) stands for the total token number routed to larger experts (4.5, 4.0, 3.0), and the sum(S) stands for the total token number routed to smaller experts (2.0, 1.0, 0.5).}
\label{token dist across ex}
\end{table}

\section{Related Work}

\subsection{FFNs Designs}
In the field of FFNs structure designs, there have been several notable works. DeepSeekMoE \cite{dai-etal-2024-deepseekmoe} introduces two strategies, namely Fine-Grained Expert Segmentation and Shared Expert Isolation. By utilizing a finer granularity of expert size, experts can focus on more specific knowledge domains. In contrast, conventional expert sizes tend to cover a wider range of knowledge. The isolated shared experts handle common knowledge across various contexts, ensuring no shared parameters among experts, thereby compressing the parameter space. To tackle the issue of Knowledge Redundancy, HyperMoE \cite{zhao2024hypermoebettermixtureexperts} also introduces HyperNetworks that contain HyperExperts to facilitate knowledge transfer between experts through conditional generation. In addition, DeLighT \cite{mehta2021delightdeeplightweighttransformer} and Apple OpenELM \cite{mckinzie2024mm1methodsanalysis} introduce block-wise scaling and layer-wise scaling, respectively. These modifications involve adjusting the width of the hidden dimension for FFNs and the number of attention heads on a per-layer basis, leading to more efficient parameter allocation and improved model performance. 

In contrast to previous works, our research focuses on the allocation of expert parameters within a single MoE layer. This approach aims to equip experts with diverse predictive capacities while ensuring load balance across computational nodes.

\subsection{Load Balance}
The LSTM MoE \cite{shazeer2017s} achieves load balance by incorporating the coefficient of variation of the load function as part of the auxiliary loss. This represents the probability of the gating network being non-zero. GShard, Switch Transformers, and ST-MoE \cite{lepikhin2021gshard, 10.5555/3586589.3586709, zoph2022stmoedesigningstabletransferable} also use a similar auxiliary load balance loss setting by introducing the average probability of each expert being routed across all tokens in the batch in the loss function. DeepSeekMoE \cite{dai-etal-2024-deepseekmoe} introduces the expert-level balance loss and the device-level balance loss to deal with the load imbalance issue caused by routing collapse. The expert-level balance loss adjusts the auxiliary loss in Switch Transformers by multiplying a coefficient to fit the different numbers of experts in DeepSeekMoE. The device-level balance loss changes the expert-level balance loss from being expert-wise to device-wise.

In our work, we utilize the balance loss from Switch Transformers \cite{10.5555/3586589.3586709}. Additionally, we propose an expert-pair allocation strategy to address the imbalance in expert sizes.

\section{Conclusion}

In this paper, we propose MoDSE, a novel structure for MoE layers. Inspired by the varying difficulties of next-token-generating tasks, we introduce the diverse size expert design, providing each expert with different prediction abilities. Our analysis of token routing distribution shows that MoDSE directs tokens to experts whose sizes are best suited for specific token generation tasks. This enhancement improves the MoE model's performance in auto-regression tasks and demonstrates superior results compared to the conventional MoE structure. Additionally, we present the expert-pair allocation method to address the issue of load imbalances in the diverse size expert design, making the MoDSE design more practical.

\section*{Limitations}

While MoDSE demonstrates superior performance, our work is subject to several limitations:
\begin{itemize}[leftmargin=*]
\item Due to limitations in computational and data resources, current experiments are conducted on small-scale MoE models, leaving the model's scalability to larger sizes unclear.
\item We obtain the aforementioned intriguing findings while training our own MoE LLM. Hence, the tokenizer and data utilized for pretraining are not available as open-source resources. We plan to apply this model design to open-source resources in our future work.
\end{itemize}

\bibliography{custom}

\appendix

\onecolumn
\section{Statistic of Tokens Routed to Each Expert}
\label{token routed}

\begin{table}[ht]
\scriptsize
\centering
\begin{tabular}{c | c c c c c c c c |  c c c}

\hline
epoch 2&2.5&2.5&2.5&2.5&2.5&2.5&2.5&2.5&max&min&max/min \\
\hline
layer0 top0&29283650&27843096&28313260&20797332&19968428&21664288&22561424&27503554&29283650 &19968428 &1.466 \\
layer0 top1&19824524&21266044&21180018&28489262&29444552&28108250&27324640&22297848&29444552 &19824524 &1.485 \\

\hline
layer1 top0&26913132&31578272&23496192&21154950&27854644&20922060&22826458&23189448&31578272 &20922060 &1.509 \\
layer1 top1&22222698&18116484&25943200&28347640&21870444&28522010&26619972&26292908&28522010 &18116484 &1.574 \\

\hline
layer2 top0&28038104&24858142&15980771&20697046&22659866&20584172&30048836&35068336&35068336 &15980771 &2.194  \\
layer2 top1&21097924&25165510&32719222&27925764&27441736&28744420&20168096&14672427&32719222 &14672427 &2.230  \\
\hline
layer3 top0&24783870&22814628&26997544&23256474&24942200&25505410&23122696&26512526&26997544 &22814628 &1.183 \\
layer3 top1&25017684&27073428&22917520&27105674&24284552&23305772&25765682&22464726&27105674 &22464726 &1.207 \\
\hline
layer4 top0&20504824&29644628&23287546&20758712&22245472&32806136&29163984&19523712&32806136 &19523712 &1.680 \\
layer4 top1&28734220&20754090&25719624&28659708&26982360&17935024&19967576&29182360&29182360 &17935024 &1.627 \\

\hline
layer5 top0&19569102&19177988&21984416&22605320&27261858&29841404&31757410&25737580&31757410 &19177988 &1.656 \\
layer5 top1&29239018&30149084&27629824&26208280&22335860&20934212&18233828&23205144&30149084 &18233828 &1.653 \\

\hline
layer6 top0&21706828&25536640&25639752&25918792&27380762&22439950&26282752&23029752&27380762 &21706828 &1.261 \\
layer6 top1&26835964&24048912&23777924&23997912&22667788&26549958&23598524&26458106&26835964 &22667788 &1.184 \\

\hline
layer7 top0&22935412&22115236&21804254&23135292&24885640&33355516&26846896&22856914&33355516 &21804254 &1.530 \\
layer7 top1&26036148&29349648&26147774&24882052&24260728&19236964&21050220&26971512&29349648 &19236964 &1.526 \\
\hline
\end{tabular}
\end{table}

\begin{table}[ht]
\scriptsize
\centering
\begin{tabular}{c | c c c c c c c c |  c c c}
\hline
epoch 3&2.5&2.5&2.5&2.5&2.5&2.5&2.5&2.5&max&min&max/min \\
\hline
layer0 top0&28840466&27370498&30216292&20019128&18389208&21972636&22685108&26016832&30216292 &18389208 &1.643  \\
layer0 top1&19648160&21190408&18650852&28781896&30648440&27082024&26291904&23216488&30648440 &18650852 &1.643  \\
\hline
layer1 top0&25124312&32307928&24555956&21069208&28992488&20350840&21091528&22018010&32307928 &20350840 &1.588 \\
layer1 top1&23455204&16594616&24372516&27749024&19862196&28540498&27946122&26990122&28540498 &16594616 &1.720 \\
\hline
layer2 top0&26050934&25051684&14964775&18084310&21737804&20938274&31295864&37386630&37386630 &14964775 &2.498 \\
layer2 top1&22030484&24361996&33255022&29913992&28102548&27683300&18286386&11876560&33255022 &11876560 &2.800 \\
\hline
layer3 top0&24287220&23189684&27490796&23700824&23515964&25244772&21362132&26718792&27490796 &21362132 &1.287 \\
layer3 top1&25154034&26214716&21516016&26275100&24873752&23033064&26755628&21688000&26755628 &21516016 &1.244 \\
\hline
layer4 top0&23147116&31119130&22808192&19547710&19158316&34343890&28756656&16629383&34343890 &16629383 &2.065 \\
layer4 top1&25875282&18430068&25610864&29503740&29509824&15279539&19579790&31721152&31721152 &15279539 &2.076 \\
\hline
layer5 top0&18971828&18990532&21079480&23400124&27013948&29703460&30990896&25359884&30990896 &18971828 &1.634 \\
layer5 top1&28894136&29686768&28161108&24903848&22162318&20255040&18251820&23195066&29686768 &18251820 &1.627 \\
\hline
layer6 top0&20292758&25357464&26143508&24792624&27778304&22373168&26254332&22518332&27778304 &20292758 &1.369 \\
layer6 top1&27724460&23550082&22739724&24591060&21575154&26145808&22859178&26324766&27724460 &21575154 &1.285 \\
\hline

layer7 top0&21360940&23354032&21291492&22724448&24471290&33274024&26836740&22197482&33274024 &21291492 &1.563 \\
layer7 top1&26940196&27452420&26136296&24861000&24253588&18341180&20699808&26825868&27452420 &18341180 &1.497 \\
\hline
\end{tabular}
\end{table}

\begin{table}[ht]
\scriptsize
\centering
\begin{tabular}{c | c c c c c c c c |  c c c}

\hline
epoch 4 &2.5&2.5&2.5&2.5&2.5&2.5&2.5&2.5&max&min&max/min \\
\hline
layer0 top0&28177806&26205268&29848370&20502658&18368704&23512666&24301080&24921296&29848370 &18368704 &1.625 \\
layer0 top1&20477908&22517138&19002768&28426338&30753764&25536248&24805638&24318394&30753764 &19002768 &1.618 \\

\hline
layer1 top0&24486260 &32865268&24768350&21399328&29389690&20514932&20790064&21623978&32865268 &20514932 &1.602 \\
layer1 top1&24175758&16210355&24277948&27526860&19510736&28411286&28305024&27419952&28411286 &16210355 &1.753 \\
\hline

layer2 top0&25388504&25202212&15168777&17673690&21703876&21467830&31400740&37832188&37832188 &15168777 &2.494 \\
layer2 top1&22868272 &24293568&33133348&30445224&28177000&27162674&18271324&11486587&33133348 &11486587 &2.885 \\

\hline
layer3 top0&24301124&23543448&27943036&23910156&23123616&25087200&20971268&26958054&27943036 &20971268 &1.332 \\
layer3 top1&25241526&25779044&21227178&26019144&25399780&23255366&27269776&21646016&27269776 &21227178 &1.285  \\
\hline
layer4 top0&24653782 &31112368&23021088&18943114&18418254&34773010&28994840&15921632&34773010 &15921632 &2.184 \\
layer4 top1&24501658&18376552&25436966&30145152&30334528&14855924&19638156&32549024&32549024 &14855924 &2.191  \\
\hline

layer5 top0&18966856&19332450&20709060&24054276&27065230&29431880&30795558&25482724&30795558 &18966856 &1.624  \\
layer5 top1&29097756 &29473352&28618584&24468104&22223408&20072908&18480540&23403384&29473352 &18480540 &1.595  \\
\hline

layer6 top0&20231524&25377478&26450518&24341292&27651192&22354528&26578036&22853380&27651192 &20231524 &1.367  \\
layer6 top1&28003118&23502284&22447872&25048596&21713974&26333100&22695000&26094096&28003118 &21713974 &1.290 \\
\hline

layer7 top0&21097412 &23748068&21571990&22617964&24818876&33164832&27007072&21811842&33164832 &21097412 &1.572  \\
layer7 top1&27303976&26917442&26225232&25201868&23978868&18211832&20889888&27108728&27303976 &18211832 &1.499  \\
\hline
\end{tabular}
\end{table}

\begin{table}[ht]
\scriptsize
\centering
\begin{tabular}{c | c c c c c c c c |  c c c}
\hline
epoch 5&2.5&2.5&2.5&2.5&2.5&2.5&2.5&2.5&max&min&max/min \\
\hline
layer0 top0&27640100&25441882&29647460&21241878&18274912&24762444&25336264&24279444&29647460 &18274912 &1.622  \\
layer0 top1&21129604&23449436&19473076&27852372&31048840&24560112&23888028&25223032&31048840 &19473076 &1.594   \\
\hline
layer1 top0&24318636&33099916&24923964&21510252&29676472&20671040&20782416&21641632&33099916 &20671040 &1.601 \\
layer1 top1&24532970 &16079311&24389856&27571050&19394450&28418796&28562626&27675378&28562626 &16079311 &1.776 \\
\hline
layer2 top0&24691036&25399028&15454448&17672260&21904492&21993528&31282512&38227050&38227050 &15454448 &2.474 \\
layer2 top1&23575500&24352420&33107372&30690030&28263692&26762368&18544144&11328960&33107372 &11328960 &2.922 \\
\hline
layer3 top0&24333048&24191826&28251336&24135400&22934528&25159592&20704260&26914344&28251336 &20704260 &1.365 \\
layer3 top1&25414176 &25530358&21180268&25910624&25791562&23370284&27649692&21777442&27649692 &21180268 &1.305 \\
\hline
layer4 top0&25475936&31272128&23376136&18702604&18103712&35131736&28930218&15631867&35131736 &15631867 &2.247 \\
layer4 top1&23939468&18453714&25330284&30688508&30782726&14711187&19640652&33077688&33077688 &14711187 &2.248 \\
\hline
layer5 top0&18863452&19334572&20713320&24381928&27191186&29867766&30909460&25362790&30909460 &18863452 &1.639 \\
layer5 top1&29341192&29495452&28880048&24284640&22366760&20127358&18637652&23491400&29495452 &18637652 &1.583 \\
\hline
layer6 top0&20187984 &25671320&26792992&24321364&27735252&22291844&26839598&22784132&27735252 &20187984 &1.374 \\
layer6 top1&28230784&23459290&22348614&25388968&21783972&26528494&22648400&26235628&28230784 &21783972 &1.296 \\
\hline

layer7 top0&21058634&23947552&21657804&22652136&24852470&33789624&27133720&21532354&33789624 &21058634 &1.605 \\
layer7 top1&27577342&26754720&26126350&25440070&23922912&18136314&21047364&27619294&27619294 &18136314 &1.523 \\

\hline
\end{tabular}
\end{table}

\begin{table}[ht]
\scriptsize
\centering
\begin{tabular}{c | c c c c c c c c |  c c c}
\hline
epoch 6&2.5&2.5&2.5&2.5&2.5&2.5&2.5&2.5&max&min&max/min \\
\hline
layer0 top0&26846484&24682704&29175398&21595708&17889516&25458556&26141992&23768864&29175398 &17889516 &1.631  \\
layer0 top1&21720052&23962840&19688954&27256824&31071132&23524024&22923076&25412416&31071132 &19688954 &1.578 \\
\hline
layer1 top0&23970390&32957156&24850036&21550080&29649160&20983372&20266464&21332620&32957156 &20266464 &1.626 \\
layer1 top1&24615920 &16029619&24296116&27293240&19129692&27855064&28678440&27661146&28678440 &16029619 &1.789  \\
\hline
layer2 top0&24160788&25256292&15526006&17374408&21852992&22145898&31014000&38229356&38229356 &15526006 &2.462 \\
layer2 top1&23802056&24234480&32763260&30754656&28037184&26299932&18572224&11095649&32763260 &11095649 &2.953  \\
\hline
layer3 top0&24031084&24316842&28304424&24218504&22555140&24962168&20376476&26794692&28304424 &20376476 &1.389  \\
layer3 top1&25410476 &25146612&20883846&25574196&25927360&23198196&27669428&21749112&27669428 &20883846 &1.325 \\
\hline
layer4 top0&25794096&31133182&23389222&18229158&17763396&35173150&28886412&15190830&35173150 &15190830 &2.315  \\
layer4 top1&23375276&18321250&25078334&30842988&30831220&14479100&19571344&33059992&33059992 &14479100 &2.283 \\
\hline
layer5 top0&18669122&19318624&20485616&24327212&27046536&29875798&30585600&25250880&30585600 &18669122 &1.638 \\
layer5 top1&29264464&29195324&28836416&23974280&22287292&19906832&18627116&23467564&29264464 &18627116 &1.571 \\
\hline
layer6 top0&19864484 &25685672&26756706&24084936&27591836&22115090&26707852&22752840&27591836 &19864484 &1.389 \\
layer6 top1&28177424&23158786&22044774&25351584&21744648&26504960&22454412&26122880&28177424 &21744648 &1.296 \\
\hline

layer7 top0&20737748&24132738&21625242&22406252&24876214&34050120&26750882&20980428&34050120 &20737748 &1.642 \\
layer7 top1&27594586&26342784&25947932&25486712&23611280&17790616&21069788&27715816&27715816 &17790616 &1.558 \\

\hline
\end{tabular}
\end{table}

\begin{table}[ht]
\scriptsize
\centering
\begin{tabular}{c | c c c c c c c c |  c c c}
\hline
epoch 7&2.5&2.5&2.5&2.5&2.5&2.5&2.5&2.5&max&min&max/min \\
\hline
layer0 top0&26519198&24218488&28953328&21836288&17812270&26098962&26883908&23335252&28953328 &17812270 &1.625  \\
layer0 top1&22122820&24452392&19954368&26980316&31297434&22854180&22180588&25815554&31297434 &19954368 &1.568   \\
\hline
layer1 top0&23771000&33043956&24818558&21663006&29826730&21030540&20202044&21301794&33043956 &20202044 &1.636 \\
layer1 top1&24850038 &15888879&24379124&27168056&19063252&27815306&28791152&27701936&28791152 &15888879 &1.812  \\
\hline
layer2 top0&23988830&25309446&15628023&17329964&21766516&22406928&30893032&38334904&38334904 &15628023 &2.453 \\
layer2 top1&24134764&24225440&32670312&30796492&28074640&26089040&18707036&10960111&32670312 &10960111 &2.981  \\
\hline
layer3 top0&23971746&24471868&28462596&24349324&22371052&24883704&20294232&26853268&28462596 &20294232 &1.402 \\
layer3 top1&25502458 &24931664&20719930&25435108&26088546&23382672&27901442&21695736&27901442 &20719930 &1.347  \\
\hline
layer4 top0&26156924&31130480&23445364&18048268&17619310&35298460&28960064&14998750&35298460 &14998750 &2.353  \\
layer4 top1&23028430&18366356&25067708&31043456&31054450&14317949&19509402&33269836&33269836 &14317949 &2.324 \\
\hline
layer5 top0&18640536&19517980&20426106&24615804&27060210&29621514&30458084&25317490&30458084 &18640536 &1.634 \\
layer5 top1&29351362&29124874&28839340&23870808&22280166&19958440&18799694&23432892&29351362 &18799694 &1.561 \\
\hline
layer6 top0&19830384 &25798716&26884716&23954710&27487748&21989570&26767928&22944108&27487748 &19830384 &1.386 \\
layer6 top1&28284866&23061998&21942802&25499504&21823700&26600504&22435532&26008718&28284866 &21823700 &1.296 \\
\hline

layer7 top0&20550378&24216228&21652904&22279728&25031802&34292704&26815672&20818174&34292704 &20550378 &1.669 \\
layer7 top1&27857508&26084144&26014746&25682830&23529792&17450062&21134866&27903840&27903840 &17450062 &1.599 \\
\hline
\end{tabular}
\caption{The statistical results in the $300M \times 8$ Baseline setting. We collected results from the 2nd to the 7th epochs, across 8 layers, for the top 2 selected experts. The value 2.5 indicates the size ratio to the input size. The ratio of the token number from the experts chosen by the most tokens to the one chosen by the least tokens varies between 1.2 and 3.0.}
\label{baseline token routing}
\end{table}
\clearpage

\begin{table}[ht]
\scriptsize
\centering
\begin{tabular}{c | c c c c c c c c |  c c c}
\hline
epoch 2&4.5&4&3&2.5&2.5&2&1&0.5&max&min&max/min \\
\hline
layer0 top0&16663346&14628973&21024906&17747456&23583046&26680430&33502182&44104732&44104732&14628973&\textbf{3.01}  \\
layer0 top1&31319104&34020424&28208976&31751180&26096736&23151636&16667122&6719973&34020424&6719973&\textbf{5.06 }\\
\hline
layer1 top0&17406648&18198192&17334890&21769320&14180341&24756332&40767830&43521696&43521696&14180341&\textbf{3.07 }\\
layer1 top1&30763940&30499152&31573372&27645132&35765450&25284810&9587062&6816535&35765450&6816535&\textbf{5.25} \\
\hline
layer2 top0&19586976&24325616&19972962&21368884&25082528&21148536&27489000&38960490&38960490&19586976&1.99 \\
layer2 top1&29188392&24619568&29219772&28441550&24441280&28347452&22529818&11147030&29219772&11147030&2.62  \\
\hline
layer3 top0&24790510&24190516&19007708&24061990&23809120&25574976&27734804&28765556&28765556&19007708&1.51  \\
layer3 top1&23839056&22489640&29992352&25760418&25223820&25166316&23124848&22338664&29992352&22338664&1.34 \\
\hline
layer4 top0&27174548&18227520&25778452&27703114&29949966&23631480&21916040&23553684&29949966&18227520&1.64  \\
layer4 top1&20633598&29841016&23545232&22325076&19747388&25945652&28058832&27838264&29841016&19747388&1.51 \\
\hline
layer5 top0&32875096&21471548&28785028&21209278&23987440&23401328&21315420&24889864&32875096&21209278&1.55 \\
layer5 top1&15750462&27894836&20562046&27668516&25120124&26419736&28766852&25752564&28766852&15750462&1.83 \\
\hline
layer6 top0&26510264&31096148&21029284&33691620&33050888&23400900&14529893&14626092&33691620&14529893&2.32\\
layer6 top1&21158036&17818752&27687472&16161102&18211424&25090944&35724830&36082348&36082348&16161102&2.23 \\
\hline

layer7 top0&23482102&25891350&28035666&25237708&27056196&28193712&21074048&18964562&28193712&18964562&1.49 \\
layer7 top1&25425668&22966264&20549536&24670248&21567910&22219540&29143860&31392108&31392108&20549536&1.53\\

\hline
\end{tabular}
\end{table}

\begin{table}[ht]
\scriptsize
\centering
\begin{tabular}{c | c c c c c c c c |  c c c}
\hline
epoch 3&4.5&4&3&2.5&2.5&2&1&0.5&max&min&max/min \\
\hline
layer0 top0&15070334&13846719&19615932&18835360&23114544&27241136&33706816&44079416&44079416&13846719&\textbf{3.18}\\
layer0 top1&32792030&34324150&29158670&30107926&25970210&21752000&15895181&5510120&34324150&5510120&\textbf{6.23}\\
\hline
layer1 top0&15652447&17078474&16281597&20633648&15702416&27111552&41650892&41399150&41650892&15652447&2.66\\
layer1 top1&32261372&31324932&32228116&28184544&33815110&22183816&7716909&7795222&33815110&7716909&\textbf{4.38}\\
\hline
layer2 top0&20773524&24170344&20823566&20389088&24895298&21765220&25850792&36842330&36842330&20389088&1.81\\
layer2 top1&27737780&24446652&27813852&28880324&24027404&27219380&23115968&12268924&28880324&12268924&2.35\\
\hline
layer3 top0&28003176&22162362&19275456&22070700&25331670&25927628&27578004&25161460&28003176&19275456&1.45\\
layer3 top1&20803522&24583920&29069288&27076532&23650454&23825340&22095564&24405668&29069288&20803522&1.40\\
\hline
layer4 top0&24875758&19542666&24944756&27656664&30196344&23059784&22450966&22783388&30196344&19542666&1.55\\
layer4 top1&22453692&27764828&23695014&21778676&18927704&26069284&26904948&27916178&27916178&18927704&1.47\\
\hline
layer5 top0&33885692&20740372&27821278&19510794&23755644&23893832&20512968&25389572&33885692&19510794&1.74\\
layer5 top1&14058357&27672176&20976378&28795244&24877254&25698686&28904132&24527914&28904132&14058357&2.06\\
\hline
layer6 top0&27157542&30087632&21600174&34075000&31486940&22504516&13894310&14704162&34075000&13894310&2.45\\
layer6 top1&20037940&18032380&26100716&15473520&19144308&25556118&35847340&35318004&35847340&15473520&2.32\\
\hline

layer7 top0&22960884&26115624&27224172&24175604&26466420&27367512&23344556&17855398&27367512&17855398&1.53\\
layer7 top1&25201832&22280822&20429580&25269804&21565168&22624818&26466540&31671918&31671918&20429580&1.55\\

\hline
\end{tabular}
\end{table}

\begin{table}[ht]
\scriptsize
\centering
\begin{tabular}{c | c c c c c c c c |  c c c}
\hline
epoch 4&4.5&4&3&2.5&2.5&2&1&0.5&max&min&max/min \\
\hline
layer0 top0&15435647&14424511&19267660&19955780&22824916&27606192&33108890&43214240&43214240&14424511&\textbf{3.00} \\
layer0 top1&32627864&33921668&29578384&29058948&26266084&21411746&16575395&6397746&33921668&6397746&\textbf{5.30}\\
\hline
layer1 top0&16210196&17697972&16761166&20585098&16109636&27657434&40953704&39862828&40953704&16109636&2.54\\
layer1 top1&31886634&30836780&31875292&28227962&33457668&21601356&8556339&9395979&33457668&8556339&\textbf{3.91}\\
\hline
layer2 top0&21672204&24096032&21556108&20462240&25123902&22063704&24916960&35946790&35946790&20462240&1.76\\
layer2 top1&26986194&24387230&27165614&28871512&23857438&27019294&24182450&13368521&28871512&13368521&2.16\\
\hline
layer3 top0&28994772&22228140&19796632&21475498&25780208&26539278&27200108&23823090&28994772&19796632&1.46\\
layer3 top1&19976316&24762380&28725508&27635016&23333142&23185024&22405388&25815018&28725508&19976316&1.44\\
\hline
layer4 top0&24433650&21353102&24672932&27729990&30954992&22579784&22541248&21572176&30954992&21353102&1.45\\
layer4 top1&23059784&26280060&24006662&21838500&18430140&26384424&26827508&29011080&29011080&18430140&1.57\\
\hline
layer5 top0&34726308&21184472&27752292&19445436&23694244&23783572&20120142&25131420&34726308&19445436&1.79\\
layer5 top1&13491518&27315996&21001964&28987466&25046010&25852086&29379572&24763488&29379572&13491518&2.18\\
\hline
layer6 top0&27976890&29705776&22562828&34037224&30612308&22620372&13733925&14588376&34037224&13733925&2.48\\
layer6 top1&19584804&18414852&25582216&15466340&19674354&25633752&36068596&35412988&36068596&15466340&2.33\\
\hline

layer7 top0&23271108&26458132&27607128&23974988&26374770&26601528&23802304&17747908&27607128&17747908&1.56\\
layer7 top1&25215760&22191064&20278950&25685524&21774488&23145478&25903480&31643244&31643244&20278950&1.56\\

\hline
\end{tabular}
\end{table}

\begin{table}[ht]
\scriptsize
\centering
\begin{tabular}{c | c c c c c c c c |  c c c}
\hline
epoch 5&4.5&4&3&2.5&2.5&2&1&0.5&max&min&max/min \\
\hline
layer0 top0&15816975&14795466&19211764&20713912&22816944&28268640&32588158&42412492&42412492&14795466&2.87\\
layer0 top1&32425422&33840080&29914948&28515312&26559608&21030924&17185960&7152014&33840080&7152014&\textbf{4.73}\\
\hline
layer1 top0&16937188&18572358&17226836&20836300&16317747&28219744&40081050&38433108&40081050&16317747&2.46\\
layer1 top1&31335500&30244788&31519254&28300250&33484212&21352168&9475388&10912962&33484212&9475388&\textbf{3.53}\\
\hline
layer2 top0&22269012&24339112&22272828&20470234&25373524&22265468&24100492&35533990&35533990&20470234&1.74\\
layer2 top1&26563754&24564608&26593544&29018298&23795360&27004666&25167544&13916547&29018298&13916547&2.09\\
\hline
layer3 top0&29869056&22211220&19869612&21278794&26273476&27061280&27096040&22964868&29869056&19869612&1.50\\
layer3 top1&19441688&24847130&28660172&28119290&23145240&22912556&22789072&26709200&28660172&19441688&1.47\\
\hline
layer4 top0&24508792&22050232&24995746&27896464&31264136&22514264&22550062&20844706&31264136&20844706&1.50\\
layer4 top1&23225490&25618340&24122212&21880170&18145540&26635318&27017360&29980044&29980044&18145540&1.65\\
\hline
layer5 top0&35033496&21404792&28073844&19516444&23752160&23848520&19989096&25006046&35033496&19516444&1.80\\
layer5 top1&13156733&27109468&21048224&29123844&25202292&26070064&29727016&25186798&29727016&13156733&2.26\\
\hline
layer6 top0&28212076&30261626&22599006&34282910&30538828&22703038&13543943&14482998&34282910&13543943&2.53\\
layer6 top1&19355896&18356678&25408908&15519957&19995936&25682232&36492304&35812544&36492304&15519957&2.35\\
\hline

layer7 top0&23494016&26171188&28317832&23879434&26252242&26539664&24251280&17718726&28317832&17718726&1.60\\
layer7 top1&25207116&22500134&19737372&25972076&22005236&23537612&25601974&32062880&32062880&19737372&1.62\\

\hline
\end{tabular}
\end{table}

\begin{table}[ht]
\scriptsize
\centering
\begin{tabular}{c | c c c c c c c c |  c c c}
\hline
epoch 6&4.5&4&3&2.5&2.5&2&1&0.5&max&min&max/min \\
\hline
layer0 top0&16093812&14975561&18904252&21328946&22726964&28518560&31707898&41303384&41303384&14975561&2.76\\
layer0 top1&31988220&33395532&29916626&27632404&26316634&20541796&17784288&7983688&33395532&7983688&\textbf{4.18}\\
\hline
layer1 top0&17366028&19162654&17884344&20815082&16328843&28165524&38934130&36902910&38934130&16328843&2.38\\
layer1 top1&30695232&29425808&30632800&28014732&33238308&21063830&10297974&12190800&33238308&10297974&\textbf{3.23}\\
\hline
layer2 top0&22550560&24133716&22691192&20426912&25260080&22143472&23408098&34945200&34945200&20426912&1.71\\
layer2 top1&26069272&24452460&25918728&28838422&23511292&26797668&25641308&14330145&28838422&14330145&2.01\\
\hline
layer3 top0&30186092&21894216&20074012&20984920&26390656&27066260&26549148&22414152&30186092&20074012&1.50\\
layer3 top1&18939456&24843012&28350640&28135678&22795104&22636304&22879156&26980088&28350640&18939456&1.50\\
\hline
layer4 top0&24114636&22645464&24971540&27665830&31490024&22283970&22382662&20005112&31490024&20005112&1.57\\
layer4 top1&23324716&24843156&23884476&21761596&17779744&26601500&26862890&30501228&30501228&17779744&1.72\\
\hline
layer5 top0&35146936&21452108&28001672&19414108&23622068&23658632&19755110&24508796&35146936&19414108&1.81\\
layer5 top1&12861604&26821580&20815188&28907844&25152160&25959822&29715866&25325488&29715866&12861604&2.31\\
\hline
layer6 top0&28441412&29818608&22813578&34151300&30129660&22669204&13298102&14237526&34151300&13298102&2.57\\
layer6 top1&19040744&18366774&25150260&15378771&19936480&25517180&36495400&35673652&36495400&15378771&2.37\\
\hline

layer7 top0&23468158&26267900&28489168&23416328&26003720&26155488&24326500&17432104&28489168&17432104&1.63\\
layer7 top1&25049572&22341700&19325668&25975586&21985580&23565524&25260574&32055100&32055100&19325668&1.66\\

\hline
\end{tabular}
\end{table}

\begin{table}[ht]
\scriptsize
\centering
\begin{tabular}{c | c c c c c c c c |  c c c}
\hline
epoch 7&4.5&4&3&2.5&2.5&2&1&0.5&max&min&max/min \\
\hline
layer0 top0&16658651&\underline{15442565}&18865092&21987256&22649968&29079684&30773936&\textbf{40200720}&40200720&15442565&2.60 \\
layer0 top1&31597378&33059774&\textbf{30060398}&27050224&26408984&19951794&18559190&\underline{8969750}&33059774&8969750&\textbf{3.69} \\
\hline
layer1 top0&18063836&20016284&18673120&20885180&\underline{16292779}&28121420&\textbf{38053416}&35551428&38053416&16292779&2.34\\
layer1 top1&30096644&28622178&29861340&27973456&\textbf{33291506}&21119694&\underline{11155827}&13537093&33291506&11155827&2.98\\
\hline
layer2 top0&22710728&24359604&23164804&\underline{20338066}&25331580&22210746&22980136&\textbf{34562164}&34562164&20338066&1.70 \\
layer2 top1&25877828&24435708&25434154&\textbf{28942052}&23496814&26774384&25992528&\underline{14703959}&28942052&14703959&1.97  \\
\hline
layer3 top0&\textbf{30709220}&21912764&\underline{20278002}&20799816&26468278&27283772&26290068&21915842&30709220&20278002&1.51 \\
layer3 top1&\underline{18606948}&24972224&28166296&\textbf{28336400}&22801280&22345112&23052706&27376696&28336400&18606948&1.52 \\
\hline
layer4 top0&24377520&23335664&25058708&27551086&\textbf{31763206}&22082748&22259234&\underline{19229592}&31763206&19229592&1.65 \\
layer4 top1&23237720&24367640&23926294&21888152&\underline{17474778}&26663284&26885158&\textbf{31214610}&31214610&17474778&1.79 \\
\hline
layer5 top0&\textbf{35550316}&21401108&28219244&\underline{19543368}&23572724&23538660&19663774&24168460&35550316&19543368&1.82 \\
layer5 top1&\underline{12647932}&26792236&20775084&28832044&25121424&26005564&\textbf{29843172}&25640084&29843172&12647932&2.36\\
\hline
layer6 top0&28698550&29936944&23038958&\textbf{34159024}&29855548&22724280&\underline{13175215}&14069042&34159024&13175215&2.59\\
layer6 top1&18882852&18490492&24931360&\underline{15474659}&20093408&25468612&\textbf{36602344}&35713964&36602344&15474659&2.37\\
\hline

layer7 top0&23527058&26429900&\textbf{28726416}&23302828&25849284&26074966&24495250&\underline{17251952}&28726416&17251952&1.67 \\
layer7 top1&25090384&22189846&\underline{19117056}&26012520&22251508&23652908&25092852&\textbf{32250640}&32250640&19117056&1.69\\
\hline
\end{tabular}
\caption{The statistical results in the $300M \times 8$ MoDSE setting. Results from the 2nd to the 7th epochs are collected, across 8 layers, for the top 2 selected experts. The values 4.5, 4, ... indicate the size ratio to the input size. Bold font in the last column indicates ratios larger than 3.00, which is the ratio of the token number from the experts chosen by the most tokens to the one chosen by the least tokens. Bold font in the middle 8 columns indicates the number of tokens from the experts chosen by the most tokens, and the underlined number is the number of tokens from the experts chosen by the least tokens}
\label{MoDSE token routing}
\end{table}

\clearpage
\section{Difficult Tokens Distribution across Experts}
\label{token dist on de}

\begin{table}[ht]
\scriptsize
\centering
\begin{tabular}{c | c c c c c c c c |  c c }
\hline
&4.5&4&3&2.5&2.5&2&1&0.5&sum of larger experts &sum of smaller experts \\
\hline
layer1 top0&208&271&324&206&127&85&93&190&-&-\\
layer1 top1&46	&159&255&122&191&135&334&262&-&-\\
layer2 top0&309&573&239&166&117&88&12&0&-&-\\
layer2 top1&248&125&429&149&131&216&187&19&-&-\\
layer3 top0&164&140&249&68&130&351&202&200&-&-\\
layer3 top1&66&	274&288&49&112&365&300&50&-&-\\
layer4 top0&211&161&150&87&378&331&144&42&-&-\\
layer4 top1&84&	44&168&117&366&287&320&118&-&-\\
layer5 top0&202&348&312&227&209&0&160&46&-&-\\
layer5 top1&110&243&142&325&155&54&280&195&-&-\\
layer6 top0&90&191&531&120&72&68&170&262&-&-\\
layer6 top1&216&198&109&149&85&124&212&411&-&-\\
layer7 top0&160&400&206&192&287&176	&44&39&-&-\\
layer7 top1&237&135&141&128&176&134	&221&332&-&-\\
layer7 top0&216&229&331&100&246&264	&48&70&-&-\\
layer7 top1&82&238&	221&127&151&199&245&241&-&-\\
\hline
top1+top2&	2649&	3729&	4095&	2332&	2933&	2877&	2972&	2477 &10473	&8326\\
top 1&	1560&	2313&	2342&	1166&	1566&	1363&	873&	849 & 6215	&3085\\
\hline
\end{tabular}
\caption{The distribution of difficult tokens across different experts.}
\label{Difficult Tokens Distribution on the Distinct Experts}
\end{table}

\end{document}